\documentclass[logo, 11pt, a4paper, onecolumn, numbering]{deepmind}
\usepackage{mathtools}
\usepackage{dsfont}
\usepackage[dvipsnames]{xcolor}
\usepackage[colorinlistoftodos]{todonotes}
\usepackage{booktabs}
\usepackage{xfrac}
\usepackage{bbm}

\usepackage{algorithm}
\usepackage{algorithmicx}

\usepackage[most]{tcolorbox}
\usepackage{xparse}
\usepackage{lipsum}
\usepackage{changepage}
\usepackage{enumitem}

\newcommand{\squishlist}{
   \begin{list}{$\bullet$}
    { \setlength{\itemsep}{0pt}      \setlength{\parsep}{3pt}
      \setlength{\topsep}{3pt}       \setlength{\partopsep}{0pt}
      \setlength{\leftmargin}{1.5em} \setlength{\labelwidth}{1em}
      \setlength{\labelsep}{0.5em} } }

\newcommand{\squishlisttwo}{
   \begin{list}{$\bullet$}
    { \setlength{\itemsep}{0pt}    \setlength{\parsep}{0pt}
      \setlength{\topsep}{0pt}     \setlength{\partopsep}{0pt}
      \setlength{\leftmargin}{2em} \setlength{\labelwidth}{1.5em}
      \setlength{\labelsep}{0.5em} } }

\newcommand{\squishend}{
    \end{list}  }

\newcommand{\two}{(2)}

\DeclareMathAlphabet{\mathpzc}{OT1}{pzc}{m}{n}

\usepackage[authoryear, sort&compress, round]{natbib} 
\usepackage{url}
\usepackage{xcolor}
\usepackage{booktabs}       
\usepackage{parskip}        
\usepackage{subcaption}     
\usepackage{makecell}       
\usepackage{wrapfig}
\usepackage{graphicx}
\usepackage[font=small]{caption}
\usepackage{bm}

\usepackage{listings}
\usepackage[utf8]{inputenc}
\usepackage{xcolor}

\definecolor{codegreen}{rgb}{0,0.6,0}
\definecolor{codegray}{rgb}{0.5,0.5,0.5}
\definecolor{codepurple}{rgb}{0.58,0,0.82}
\definecolor{backcolour}{rgb}{0.95,0.95,0.92}

\lstdefinestyle{mystyle}{
    backgroundcolor=\color{backcolour},   
    commentstyle=\color{codegreen},
    keywordstyle=\color{magenta},
    numberstyle=\tiny\color{codegray},
    stringstyle=\color{codepurple},
    basicstyle=\ttfamily\footnotesize,
    breakatwhitespace=false,         
    breaklines=true,                 
    captionpos=b,                    
    keepspaces=true,                 
    numbers=left,                    
    numbersep=5pt,                  
    showspaces=false,                
    showstringspaces=false,
    showtabs=false,                  
    tabsize=2
}

\graphicspath{{figures/}}

\title{Hyperparameter Selection for Offline Reinforcement Learning}

\correspondingauthor{tpaine@google.com, paduraru@google.com}

\reportnumber{001} 

\definecolor{plot_red}{HTML}{e53935}
\definecolor{plot_green}{HTML}{44a047}
\definecolor{plot_blue}{HTML}{1e88e5}

\author[*,1]{Tom Le Paine}
\author[*,1]{Cosmin Paduraru}
\author[1]{Andrea Michi}
\author[1]{Caglar Gulcehre}
\author[1]{Konrad \.Zo\l{}na}
\author[1]{Alexander Novikov}
\author[2]{Ziyu Wang}
\author[1]{Nando de Freitas}

\affil[*]{Equal contributions}
\affil[1]{DeepMind}
\affil[2]{Google}

\begin{abstract}
Offline reinforcement learning (RL purely from logged data) is an important avenue for deploying RL techniques in real-world scenarios.
However, existing hyperparameter selection methods for offline RL break the offline assumption by evaluating policies corresponding to each hyperparameter setting in the environment.
This online execution is often infeasible and hence undermines the main aim of offline RL.
Therefore, in this work, we focus on \textit{offline hyperparameter selection}, i.e. methods for choosing the best policy from a set of many policies trained using different hyperparameters, given only logged data.
Through large-scale empirical evaluation we show that:
1) offline RL algorithms are not robust to hyperparameter choices,
2) factors such as the offline RL algorithm and method for estimating Q values can have a big impact on hyperparameter selection,
and 3) when we control those factors carefully, we can reliably rank policies across hyperparameter choices, and therefore choose policies which are close to the best policy in the set.
Overall, our results present an optimistic view that offline hyperparameter selection is within reach, even in challenging tasks with pixel observations, high dimensional action spaces, and long horizon.

\end{abstract}

\begin{document}
\maketitle

\section{Introduction}

The desire to apply reinforcement learning methods to a broad range of real-world problems has prompted renewed interest in \emph{offline reinforcement learning} (ORL), i.e. methods that can learn a policy from logged data \citep{fujimoto2018off,levine2020offline}. These methods are useful when it is challenging, risky or expensive to have an arbitrary policy interact with an environment, for example for robotics \citep{cabi2019framework}, self-driving cars, health care \citep{futoma2020popcorn}, or dialogue \citep{jaques2019way}.

Despite a range of impressive results, the quality of different ORL algorithms depends heavily on hyperparameter choice. Of course, this is a known issue in machine learning in general and in reinforcement learning in particular (\citet{henderson2018deep}; see Section \ref{sec:background} for background). The standard way of performing hyperparameter selection in RL is to evaluate policies online by interacting with the environment. However, picking hyperparameters for offline RL methods requires offline methods, where direct interaction with the environment is not allowed (as pointed out, for instance, by \citet{wu2019behavior}). Therefore, we must devise offline statistics to rank policies produced by different hyperparameter settings. 

\begin{wrapfigure}{r}{0.4\linewidth}
\vspace{0.2cm}
    \includegraphics[width=\linewidth]{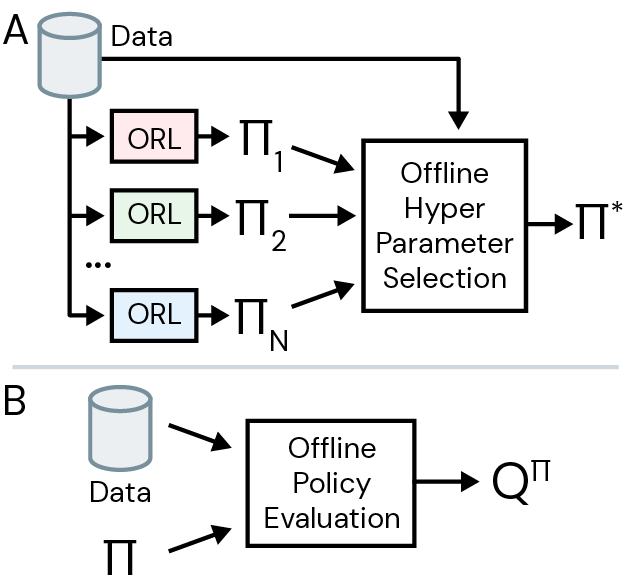}
    \caption{\textbf{Offline hyperparameter selection compared to offline policy evaluation.} (A) In offline hyperparameter selection, we learn a set of $N$ policies using offline RL with different hyperparameters, and attempt to pick the best policy $\pi^{*}$. For both learning and hyperparameter selection, we have access to the same logged data, and \emph{not} to online interactions with the environment. (B) Offline Policy Evaluation is a closely related problem where the goal is to estimate the value of a policy given data.}
    \label{fig:offline_hyperparameter_tuning}
    \vspace{-2cm}
\end{wrapfigure}

This paper presents a thorough empirical study of offline hyperparameter selection for offline RL. It contrasts different approaches for selecting the best out of $N$ hyperparameter settings purely from logged data, as shown in Figure~\ref{fig:offline_hyperparameter_tuning}. In particular, this study:
\begin{itemize}
    \item Uses simple and scalable evaluation metrics to assess the merit of different approaches for offline RL  hyperparameter selection. 
    \item Uses challenging domains which require high-dimensional action spaces, high dimensional observation spaces, and long time horizons.
    \item Focuses on common important hyperparameters: model architecture, optimizer parameters, and loss function.
\end{itemize}

Our experiments confirm that ORL algorithms are not ro-
bust to hyperparameter choices, strengthening the case for
developing reliable offline hyperparameter selection methods. Additionally, through our experiments we identify three important choices that can affect how effective offline hyperparameter selection can be:
\begin{itemize}
    \item The choice of offline RL algorithm: We find that algorithms that encourage policies to stay close to the behavior policy are easier to evaluate and rank. 
    \item The choice of Q estimator: We find Q values estimated by the OPE algorithm we use, Fitted Q-Evaluation, to be more accurate than ORL estimates. 
    \item The choice of statistic for summarizing the quality of a policy: the average critic value of the initial states works better than alternatives.
\end{itemize}

When combined, these result in a strong strategy for offline hyperparameter selection across several challenging tasks.

\section{Offline Hyperparameter Selection}

An essential goal of offline RL is to enable the application of reinforcement learning methods in real-world scenarios where only logged data can be used. Therefore, hyperparameter selection for offline RL should follow the same assumption (see Figure \ref{fig:offline_hyperparameter_tuning}).

\begin{wrapfigure}{r}{0.4\linewidth}
\vspace{0.1cm}
    \includegraphics[width=\linewidth]{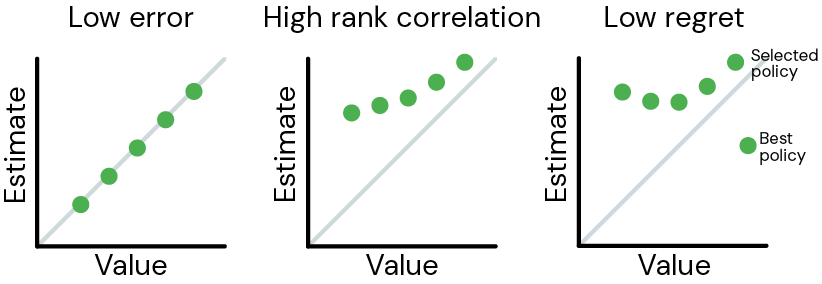}\hfil
    \caption{\textbf{Evaluating performance of offline hyperparameter selection.} Each point represents a policy in the set, plotted according to its actual value and estimated value. OPE aims to minimizing the error between the value and the estimate (moving points closer to the diagonal). In contrast, in OHS it is sufficient to (i) rank the policies as measured by spearman rank correlation, or (ii) select a policy whose value is close to the value of the best policy (i.e. has low regret).}
    \label{fig:metrics}
    \vspace{-0.2cm}
\end{wrapfigure}
Offline hyperparameter selection (OHS) is closely related to offline policy evaluation (OPE), which focuses on estimating a value function based on offline data.\footnote{Offline policy evaluation is in turn very closely related to off-policy evaluation. Both estimate the value function of an evaluation policy using data produced by a different policy, typically called the behavior policy, but off-policy evaluation can be performed online. In this paper, since we are only concerned with the offline setting, they are equivalent and we use the same acronym (OPE) for both.} There are two salient differences between OHS and OPE. First, in OHS there is a known relationship between the data and the policies which may be leveraged to simplify the problem, whereas in OPE this relationship is generally unknown. Second, in OHS we mainly care about picking the best policy (or close to the best) from a set, not precisely assessing the quality of a policy as OPE aims to do. As a result, OHS and OPE may focus on optimizing different performance metrics (see Figure \ref{fig:metrics}).

This paper is concerned with hyperparameter \emph{selection}, which is subtly different from hyperparameter tuning. Hyperparameter selection refers to picking the best out of a set of given policies that were trained with different hyperparameters, whereas tuning includes both selection and a strategy for searching the hyperparameter space.

We use standard RL nomenclature \citep{sutton1998}. In short, the agent policy takes $s$ as input states to generate actions $a$ which yields rewards $r$ that are discounted using $\gamma$. They all can be subscripted with a discrete timestep $t \geq 0$ and batched. The policy $\pi_\phi$ and critic $Q_\theta$ are parametrized by $\phi$ and $\theta$, respectively.

\subsection{Offline Statistics for Policy Ranking}
\label{sec:statistics}

As discussed in the Introduction, the challenge when performing offline hyperparameter selection is to rank several policies using statistics computed solely from offline data. We envision the following workflow to apply offline hyperparameter selection in practice:
\begin{enumerate}
    \item Train offline RL policies using several different hyperparameter settings.
    \item For each policy, compute scalar statistics summarizing the policy's performance (without interacting with the environment).
    \item Pick the top $k$ best policies according to the summary statistics to execute in the real environment.
\end{enumerate}

All of the statistics considered in this paper will be based on the estimated value functions (also referred to as "critic") for policies trained by offline RL methods with different hyperparameter settings. We can obtain these critics from one of two sources:
\begin{itemize}
    \item \textbf{ORL} \hspace{0.2cm} We can simply use the critic learned during offline RL training for that hyperparameter setting. For methods which do not usually leverage a critic, we additionally train a critic which does not impact the training but can be used to obtain statistics - see Section \ref{orl_algos} for a list of the ORL methods we use.
    \item \textbf{OPE} \hspace{0.2cm} We use Fitted Q Evaluation (FQE, see Section \ref{ope_algos}) to retrain a critic for the policy generated by the offline RL algorithm.
\end{itemize}

We then compute scalar values for the purpose of ranking policies by calculating a statistic based on the critic $Q_\theta$ and the dataset ${\mathcal D}$. We do this in one of two ways:
\begin{itemize}
    \item $\bm{\hat V(s_0)}$ \hspace{0.2cm} We use an estimate of the expected value of the evaluation policy for the initial state distribution $\mathbb{E}_{s_0\sim {\mathcal D}}[Q_\theta(s_0, \pi_e(a)]$. This is an estimate of what we care about, i.e. the value we would achieve by running the policy in the environment from initial states.
    \item \textbf{Soft OPC} \hspace{0.2cm} We use the soft off-policy classification (OPC) statistic proposed in \citet{irpan2019off}. This statistic can be written as $\mathbb{E}_{(s, a)\sim {\mathcal D}, success}[Q_\theta(s, a)] - \mathbb{E}_{(s, a)\sim {\mathcal D}}[Q_\theta(s, a)]$. Here success indicates whether $(s,a)$ is part of a successful trajectory, namely one whose return is above a certain threshold; we try different values for this threshold in our experiments and pick the best one.
\end{itemize}

We also considered two additional statistics: the average Q across all states, which performed similarly but slightly worse than $\hat V(s_0)$, as well as the average TD error, which clearly underperformed the main statistics considered. We defer these additional statistics to Appendix \ref{app:ranking_results}.

\subsection{Metrics for Evaluating Offline Hyperparameter Selection}
\label{sec:metrics}
As discussed in the previous section, we are interested in selecting good hyperparameter settings based on summary statistics for the corresponding ORL policies. Therefore, we use the evaluation metrics that aim to capture how useful different statistics are for ranking multiple policies and selecting the best one(s). 

In order to compute the metrics, we first obtain a reliable estimate of the actual expected discounted return for each policy when starting in the environment's initial state distribution, by running that policy in the actual environment. This estimate will be used as the ground truth, and will be referred to as the \textbf{actual value}. We then compute the following metrics:
\begin{itemize}
\item {\bf Spearman's rank correlation}: First compute the rank values of the different policies according to both the summary statistics and the actual values. Spearman's rank correlation is the Pearson correlation between the two sets of rank values.
\item {\bf Regret @ k}: First compute the top-k set, i.e. the k policies with the highest summary statistic values. Regret @ k is the difference between the actual value of the best policy in the entire set, and the actual value of the best policy in the top-k set. This metric aims to answer the question "If we were able to run policies corresponding to k hyperparameter settings in the actual environment and get reliable estimates for their values that way, how far would the best in the set we picked be from the best of all hyperparameter settings considered?".
\item {\bf Absolute error}: The absolute value of the difference between the statistic $\hat V(s_0)$ and the actual values. This does not measure ranking quality directly, but we include it here because zero absolute error would correspond to perfect ranking, and because it is a standard measure in the OPE literature.
\end{itemize}

\section{Experimental Setup}

In the previous section, we described offline hyperparameter selection, and the statistics and metrics we consider. In this section we describe additional elements of the experimental setup including the tasks, algorithms, and hyperparameters considered.

\subsection{Tasks}
\begin{figure}[t!]
    \centering
    \includegraphics[width=\linewidth]{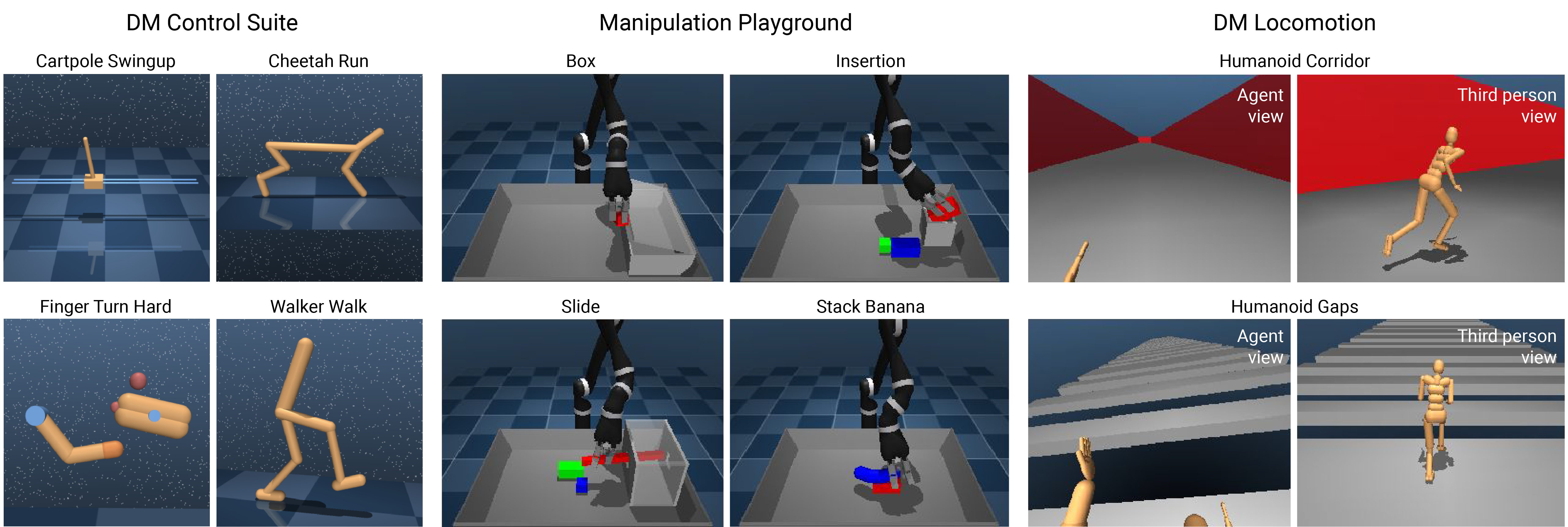}\hfil
    \caption{\textbf{Tasks considered.} We consider tasks from three continuous control domains: 1) DM Control Suite involves low dimensional action spaces from features of the MDP state, 2) Manipulation playground involves low dimensional action spaces from proprioceptive features of a Jaco arm, as well as visual representation of the scene, 3) DM Locomotion involves control of a high action space humanoid avatar, from visuals provided by an egocentric camera controlled by the policy.}
    \label{fig:offline_policy_selection}
\end{figure}
Real-world problems are diverse, spanning a range of challenging properties. We want to verify offline hyperparameter selection for high dimentional action space, observation space, long time horizon problems, including hard exploration problems. To achieve this end, we use ten tasks spanning two domains included in RL Unplugged \citep{gulcehre2020rl}, as well as one robotic manipulation domain \citep{wang2020critic}. In Figure \ref{fig:offline_policy_selection}, we illustrate some of those tasks and below we describe them in detail. For additional details about the datasets we used see Appendix~\ref{app:dataset}.

\textbf{DM Control Suite} \hspace{0.2cm}
A set of continuous control tasks introduced by \citep{tassa2018deepmind} implemented in MuJoCo \citep{todorov2012mujoco}. We choose four tasks with various levels of difficulty from the subset included in \citet{gulcehre2020rl}. For all tasks in this domain we use a feature representation of the MDP state, including proprioceptive information such as joint positions and velocities.

\textbf{Manipulation tasks} \hspace{0.2cm}
These tasks require continuous control of a Kinova Jaco robotic arm with 9 degrees of freedom (simulated in MuJoCo \citep{todorov2012mujoco}). The tasks include manipluation problems such as picking up a block and placing it in a box. We use joint velocity control (at 20HZ) of all 6 arm joints and the 3 joints of the hand. The agent observes the proprioceptive features directly, but can only infer the objects on the table from pixel observations. Two camera views of size 64 × 64 are provided: one frontal camera covering the whole scene, and an in-hand camera for closeup of the objects. The episodes are of length 400 and the reward function is binary depending on whether the task is successfully executed.

\textbf{DM Locomotion} \hspace{0.2cm}
A set of continuous control tasks which involve controlling a 56 degrees of freedom humanoid avatar, resulting in a large action spaces \citep{tassa2020dm_control}. For all tasks, we learn directly from large observation spaces (i.e. 64x64 RGB images). These images are generated by an egocentric camera under the control of the policy. We focus on humanoid corridor and humanoid gaps, which are difficult tasks but do not require long-term memory.

\subsection{Offline RL Algorithms}
\label{orl_algos}

Our experiments select among policies produced using different hyperparameter settings for three offline reinforcement learning algorithms listed below.

\textbf{Behavior Cloning (BC; \cite{pomerleau1989alvinn})} \hspace{0.2cm}
BC's policy objective attempts to match the actions from the behavior data. Standard behavior cloning does not train a value function, but we do train a value function alongside the policy to enable downstream policy evaluation. We note that the training procedure is agnostic to the critic trained.

\textbf{Critic Regularized Regression (CRR; \citet{wang2020critic})} \hspace{0.2cm}
The policy objective of CRR attempts to match the actions from the behavior data, while also preferring actions with high value estimates. This encourages the policy to be close to the behavior policy for some, but not all states.

\textbf{Distributed Distributional Deep Deterministic Policy Gradient (D4PG; \cite{barth2018distributed})} \hspace{0.2cm}
D4PG's policy objective directly optimizes the critic. As a result, there is no regularization towards the behavior policy.

\begin{wrapfigure}{r}{0.475\linewidth}
\vspace{-0.1cm}
\centering
\refstepcounter{table}
\caption*{Table \thetable~~| \textbf{Policy updates for considered algorithms.} BC and CRR regress to actions logged in the data while D4PG relies on critic estimates.
\label{tab:policy_update}
}
\small
\begin{tabular}{lc}
\toprule
\thead{Algorithm} & \thead{Policy Update} \\
\midrule
BC  & $-\nabla_{\phi}\log \pi_{\phi} (a_t | s_t) $\\
CRR          & $-\nabla_{\phi}\log \pi_{\phi} (a_t | s_t) \cdot w^*$ \\
D4PG         & $\nabla_{\phi} Q_\theta(s_t, \pi_\phi(s_t))$ \\
\midrule
FQE          &   None \\
\bottomrule
\multicolumn{2}{r}{{\em $^*w$ reflects CRR weighting as in \citep{wang2020critic}}}\\
\end{tabular}
\vspace{-0.5cm}
\end{wrapfigure}
By design, all the algorithms share the same value objective.
However, they use the same value function update, which is
\begin{equation}\label{eg:critic_update}
    - \nabla_\theta d(Q_\theta(s_t, a_t), r_t + \gamma \mathbb{E}_{a \sim \pi_\phi(s_{t+1})} Q_\theta (s_{t+1}, a) ),
\end{equation}
where $d$ is a divergence measure.
As the value update is the same for all the algorithms, they are defined by their policy updates which are shown in Table~\ref{tab:policy_update}.

Since the only difference between the algorithms is the policy objective, we can isolate its effect on downstream policy evaluation.

Finally, we want to reiterate that two of the algorithms, BC and CRR, encourage the policy to stay close to the behavior policy, whereas D4PG does not, allowing the policy to freely optimize against the critic. As shown in our experiments (see Section~\ref{sec:results}), this characteristic strongly influences the OHS accuracy.

\subsection{Offline Policy Evaluation Algorithms}
\label{ope_algos}

We re-evaluate the policies generated by the ORL algorithms described above using FQE.

\begin{wrapfigure}{r}{0.5\linewidth}
\vspace{-0.7cm}
\begin{minipage}{1\linewidth}
\begin{algorithm}[H]
\small
\caption{\small Fitted Q Evaluation \label{alg:fqe}}

\textbf{Input:} Dataset ${\cal D}$, policy $\pi_e$ to evaluate

\textbf{For} $n_{updates}$ do
 
\hspace{1cm} Sample $\{s_i, a_i, r_i, s'_i\}_{i=1}^{batch\_size}$ from ${\cal D}$
      
\hspace{1cm} Update critic according to Eq.~\ref{eg:critic_update}
      
\end{algorithm}
\end{minipage}
\vspace{-0.4cm}
\end{wrapfigure}
\textbf{Fitted Q Evaluation (FQE; \cite{le2019batch})} \hspace{0.2cm}
Our FQE algorithm employs the same value function updates as the above ORL methods but keeps the policy fixed. Pseudocode for it can be found in Algorithm~\ref{alg:fqe}, and a simplified version of the code we use can be found in Appendix \ref{sec:fqe_code}.

FQE was shown to work well in a recent suite of experiments on relatively simple problems \citep{voloshin2019empirical}. Using it allows us to interrogate how much re-evaluation with the same objective and dataset improves policy evaluation. In our experiments, we assume the offline RL algorithms and FQE have access to the same data.

There are many possible OPE algorithms we could try, as discussed in Section \ref{sec:background}. We focus on Fitted Q Evaluation due to its simplicity and scalability. Other OPE methods have to solve difficult estimation problems, such as learning a transition model from pixels or computing importance sampling corrections for continuous actions. Fitted Q Evaluation foregoes estimating these complex quantities by directly estimating the value function of the policy being evaluated.

\subsection{Hyperparameters and Other Implementation Details}

\begin{wrapfigure}{r}{0.4\linewidth}
\vspace{-0.1cm}
\centering
\refstepcounter{table}
\caption*{Table \thetable~~| \textbf{Hyperparameters considered.} We consider parameters specific to the model architecture, the optimizer, and loss function.\label{tab:hyperparameters}
}
\small
\begin{tabular}{ll}
\toprule
\thead{Hyperparameter} & \thead{Values}\\
\midrule
Hidden size & 64, 1024 \\
Num blocks & 1, 5  \\
Learning rate & 0.001, 0.00001 \\
Learner steps & range(50k, 250k, 25k) \\
Algorithms & BC, CRR, D4PG \\
Beta (for CRR) & 0.1, 10 \\
\bottomrule
\end{tabular}
\vspace{-0.2cm}
\end{wrapfigure}
For each task, we train policies using the hyperparameters described in Table \ref{tab:hyperparameters}, resulting in 256 policies per task (64 BC, 128 CRR, 64 D4PG). We considered hyperparameters that affect the model architecture (hidden size, num blocks\footnote{The CRR paper \citep{wang2020critic} uses blocks composed of two linear layers followed by layer norm and residual connection, and we used the same architecture.}), the optimizer (learning rate and learner steps), and loss function (algorithm and loss term beta), since these are known to be important for many machine learning problems. The search space in grid search grows rapidly with respect to the number of hyperparameters considered and unique hyperparameter settings. Thus, we tried to choose a small representative set that is reasonably broad in terms of performance. Our implementations are based on open-sourced D4PG implementation from Acme~\citep{hoffman2020acme}.

Each policy comes with an associated critic $Q_\theta$, which we use to calculate ORL statistics. In addition, we re-evaluate each policy and hyperparameter choice using FQE for each combination of the hyperparameters in Table \ref{tab:hyperparameters}, meaning that we run the FQE algorithm 256 times per task. We use these resulting critics to generate the OPE statistics. Finally, to obtain the actual value, we run each policy in the environment for 100 episodes. In the experiments we compare the actual values to the ORL and OPE statistics.

In this paper, we use distributional critics \citep{dabney18a,BellemareDM17} for all our algorithms, including FQE. This means that the value function $Q_\theta$ in Eq.~\ref{eg:critic_update} is represented as a discrete distribution, and the discrepancy measure $d$ is the cross-entropy between the two distributions, as in \citet{barth2018distributed} and \citet{wang2020critic}. In Appendix \ref{app:fqe_without_distributional} we compare FQE estimates with and without a distributional critic and find they are similar.

\section{Results}\label{sec:results}

Our experiments confirm that hyperparameter choice does play an important role in the performance of the ORL algorithms we use on our set of tasks (see for example the range of actual values (x axis) on the graphs in Figure 4). The results presented in this section aim to shed light on the conditions under which the statistics in Section \ref{sec:statistics} rank these hyperparameter choices well.

\subsection{Overestimation}
\begin{figure}[t!]
\includegraphics[width=\linewidth]{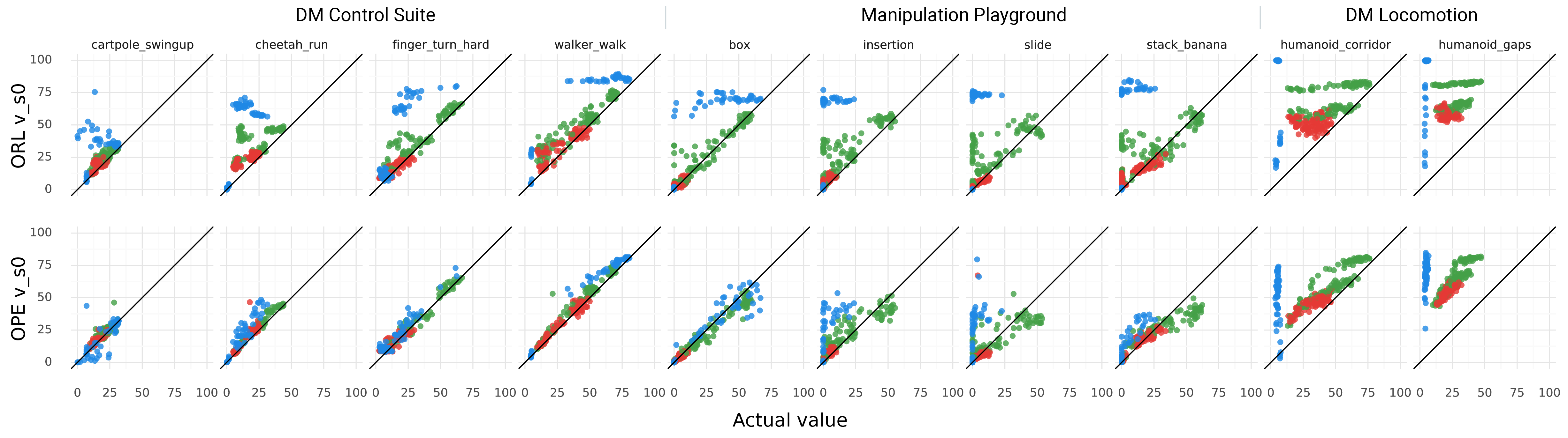}
\caption{\textbf{Value estimates vs actual values.} Each point represents a policy trained using different hyperparameters including different algorithms (\textcolor{plot_red}{$\blacksquare$} BC, \textcolor{plot_green}{$\blacksquare$} CRR, \textcolor{plot_blue}{$\blacksquare$} D4PG). (top) Value estimates from offline RL algorithms. Notice nearly all values are over-estimated (i.e. lay above the diagonal) to some degree. D4PG over-estimates the most, followed by CRR, then BC. (bottom) Value estimates from re-evaluating policies using offline policy evaluation, specifically FQE. Re-evaluation significantly reduces over-estimating in these domains, though D4PG deviates the most from the actual values.}
\label{fig:estimates_vs_values}
\end{figure}
\begin{wrapfigure}{r}{0.4\linewidth}
\vspace{-0.9cm}
\includegraphics[width=1\linewidth]{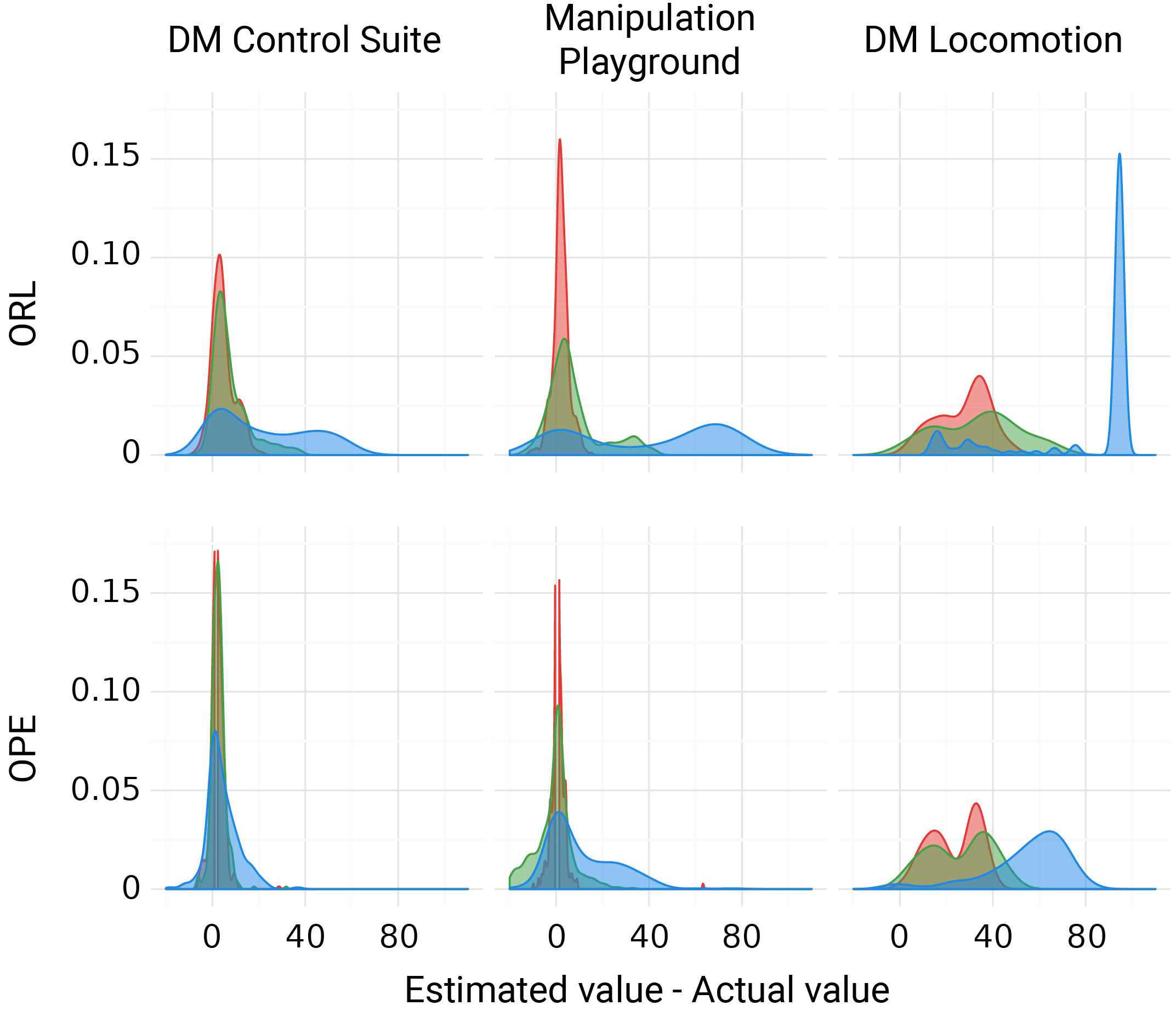}
\caption{\textbf{Distribution of over-estimation.} Summary of the distribution of over-estimation for each algorithm (\textcolor{plot_red}{$\blacksquare$} BC, \textcolor{plot_green}{$\blacksquare$} CRR, \textcolor{plot_blue}{$\blacksquare$} D4PG) and task domain. Re-evaluation by OPE (bottom plot) significantly reduces over-estimation across all domains and algorithms.}
\label{fig:over_estimation}
\vspace{-0.6cm}
\end{wrapfigure}
In the top row of Figure~\ref{fig:estimates_vs_values} and~\ref{fig:over_estimation} we compare the actual values against the ORL $\hat V(s_0)$ statistics. We show that, on all tasks, ORL's $\hat V(s_0)$ overestimates the value. In extreme cases (e.g. humanoid environments; see Figure~\ref{fig:estimates_vs_values} (top)), according to the ORL statistics, D4PG deceptively looks to produce the best policy, while actually being the worst.

We find a clear over-estimation trend -- statistics tend to over-estimate the most on D4PG, followed by CRR, followed by BC. Again, we note that BC and CRR attempt to produce policies that are similar to the behavior policy, whereas D4PG does not. This may make it easier to estimate the value of the policies they produce, given only the behavior data.
In terms of task domains, statistics tend to over-estimate the most on DM Locomotion, followed by Manipulation Playground, followed by DM Control Suite.

In the bottom row of Figure~\ref{fig:estimates_vs_values} and~\ref{fig:over_estimation}, we compare the actual values against the OPE $\hat V(s_0)$ statistics. We show that re-evaluation using FQE significantly reduces over-estimation across all algorithms and tasks. Unfortunately, the over-estimation trends described above remain. Specifically, for D4PG on DM Locomotion, the statistic still over-estimates significantly.

Finally, the plots in Figure \ref{fig:estimates_vs_values} suggest better performing policies overestimate less. This finding highlights the importance of reducing overestimation for offline RL to learn better policies.

\subsection{Ranking Quality}
\begin{figure}[t!]
\includegraphics[width=\linewidth]{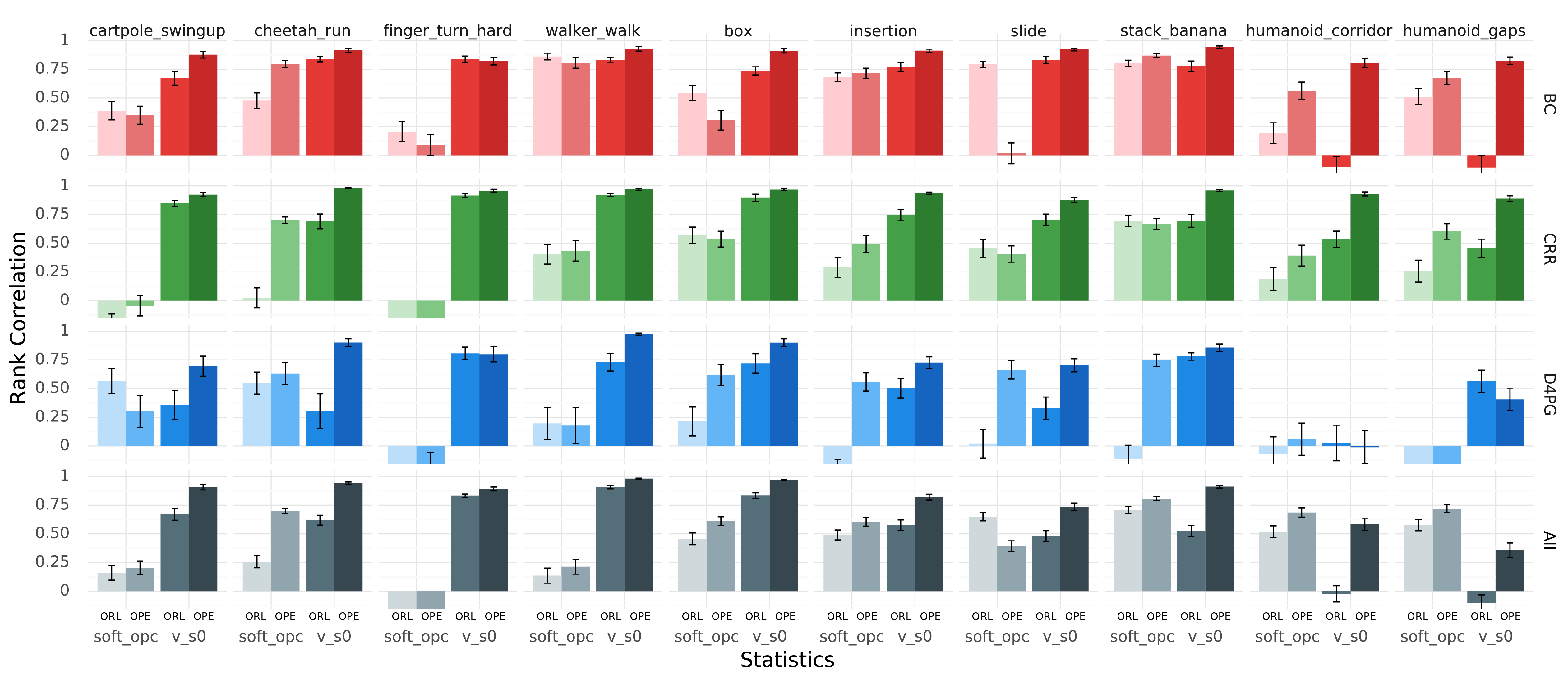}
\caption{\textbf{Rank correlation within algorithm and across all algorithms.} We compare the rank correlation between the actual value and two policy statistics: $\hat V(s_0)$ and Soft OPC (see additional statistics in Appendix \ref{app:ranking_results}). We use both ORL and OPE critics. In general in terms of ranking, OPE $\hat V(s_0)$ performed best, followed by ORL $\hat V(s_0)$, followed by OPE Soft OPC, followed by ORL Soft OPC. Also note that CRR followed by FQE re-evaluation works best across all domains.}
\label{fig:rank_correlation}
\end{figure}
\begin{figure}[t!]
\includegraphics[width=\linewidth]{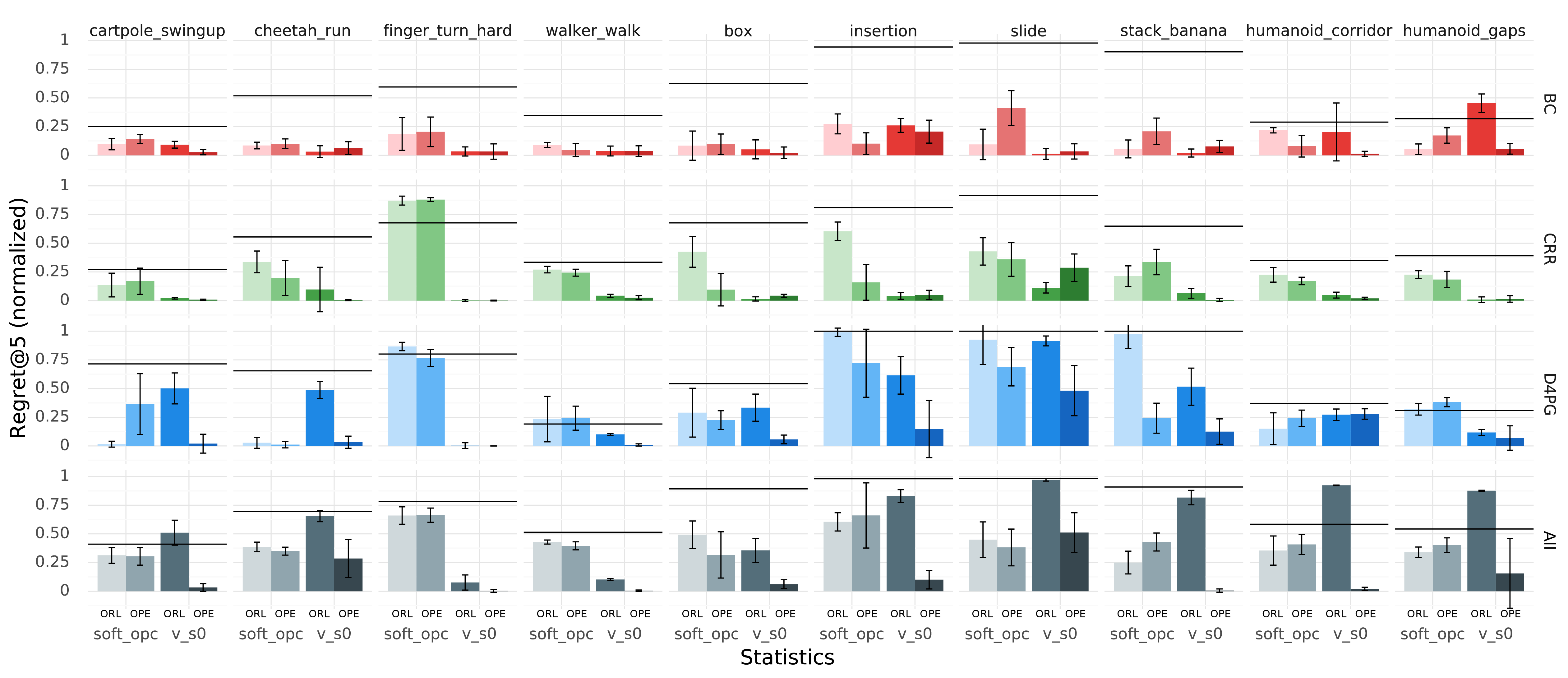}
\caption{\textbf{Regret@5 within algorithm and across all algorithms.}
We compare the normalized regret@5 between the actual value and various policy statistics from ORL and OPE critics ($\hat V(s_0)$, Soft OPC, additional statistics in Appendix \ref{app:ranking_results}). We normalize by the value of the best policy. The black horizontal line indicates the regret associated with the policy of median value. The regret follows similar trends to rank correlation. Note especially that that BC and CRR have generally low regret across all tasks, usually much lower than the regret associated with picking the policy with median value, and often close to zero.}
\label{fig:regret_at_5}
\end{figure}
In Figure~\ref{fig:rank_correlation} we compute the rank correlation between the actual values and various statistics, broken down by ORL algorithm, and across all algorithms. The ORL $\hat V(s_0)$ statistic has better rank correlation for BC and CRR (algorithms that encourage the policy to stay close to the behavior policy) than for D4PG. The ORL $\hat V(s_0)$ statistic also performs better on environments which showed less overestimation like DM Control Suite, and worse on environments with more overestimation like DM Locomotion. For the hardest combination, D4PG on the DM Locomotion tasks, the rank correlation is quite low. Because D4PG is hard to rank in this setting, it is also hard to rank policies across all algorithms.

The same general trends hold for the OPE $\hat V(s_0)$ statistic, but it has higher correlation than ORL $\hat V(s_0)$ across the board. Especially for BC and CRR on the DM Locomotion tasks. For BC and CRR, OPE $\hat V(s_0)$ rank correlation is above 0.9 for most tasks. But for D4PG and across all algorithms on DM Locomotion is still quite low.

Similar trends for ORL and OPE $\hat V(s_0)$ statistics can be seen in the regret plots in Figure \ref{fig:regret_at_5}. BC and CRR have generally low regret across all tasks, usually much lower than the regret associated with picking the policy with median value, and often close to zero. Regret for D4PG is often low, and usually better than the median choice. For D4PG the regret for OPE $\hat V(s_0)$ tends to be lower than for ORL $\hat V(s_0)$.

Figures \ref{fig:rank_correlation} and \ref{fig:regret_at_5} indicate that ranking based on $\hat V(s_0)$ is preferable to ranking based on the Soft OPC statistic in terms of both ranking correlation and regret. For instance, OPE $\hat V(s_0)$ has higher rank correlation than OPE Soft OPC for nearly all tasks. It is not clear why this is the case. One potential explanation is Soft OPC depends on the value estimates along unsuccessful trajectories. It is possible the learned policies deviate from the behavior policy greatly on these states, and the value estimates end up being poor, making the difference in value between successful vs unsuccessful states less meaningful.

\subsection{FQE Sensitivity to Its Own Hyperparameters}

\begin{wrapfigure}{r}{0.45\linewidth}
\vspace{-0.5cm}
\includegraphics[width=\linewidth]{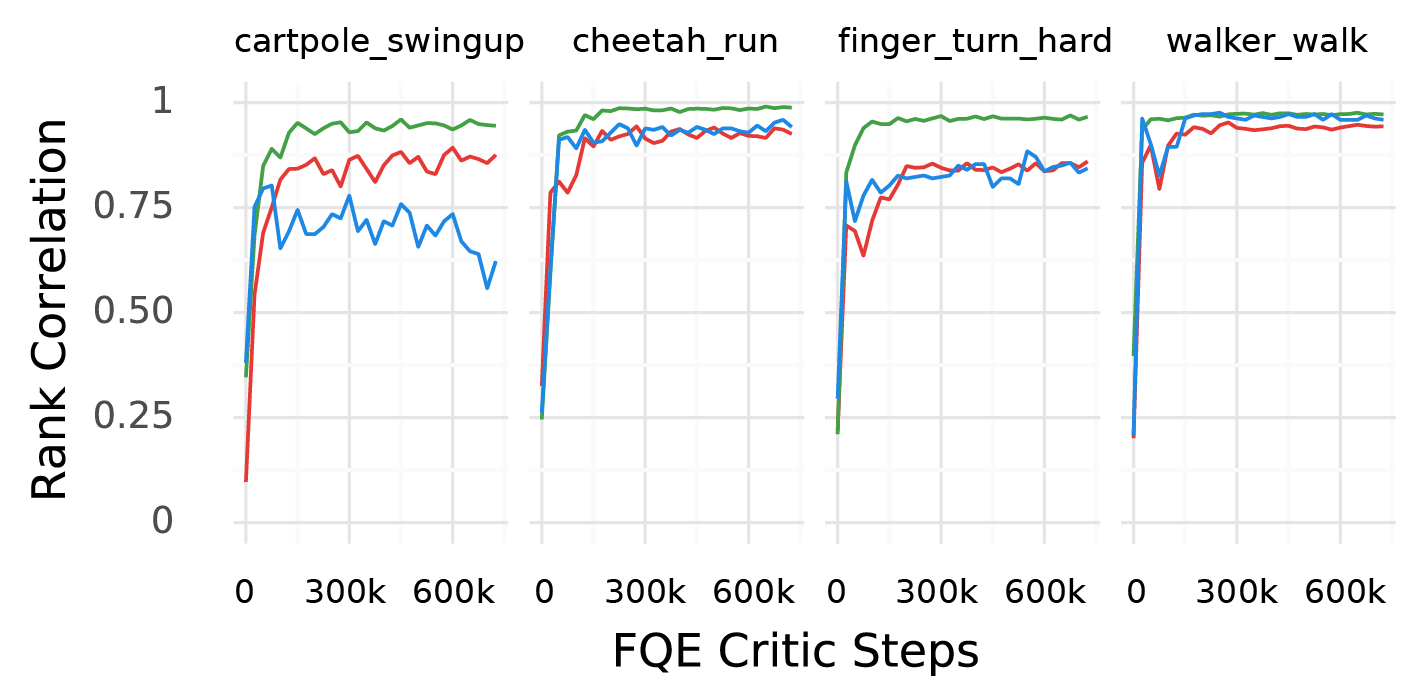}
\caption{\textbf{Rank correlation vs FQE learner steps.} (\textcolor{plot_red}{$\blacksquare$} BC, \textcolor{plot_green}{$\blacksquare$} CRR, \textcolor{plot_blue}{$\blacksquare$} D4PG) FQE is relatively stable to the number of learner steps. Though D4PG on cartpole\_swingup eventually diverges.}
\label{fig:rank_vs_steps}
\vspace{-0.2cm}
\end{wrapfigure}
Our FQE implementation was based on an existing CRR implementation and we used its default hyperparameters for all tasks (hidden size = 1024, num blocks = 4, learning rate = 0.0001, learner steps = 250k). We did not perform an exhaustive investigation of FQE's sensitivity to hyperparameters, but we did look into its sensitivity to the number of learner steps on the control suite. Figure \ref{fig:rank_vs_steps} suggests that FQE is not very sensitive to the number of learner steps and, in general, does not seem to get worse as it runs for longer. Finding a good way to tune FQE hyperparameters remains an open problem for future research.

\section{Related Work}
\label{sec:background}

Early examples of offline/batch RL include least-squares temporal difference methods~\citep{bradtke1996linear,lagoudakis2003least} and fitted Q iteration~\citep{ernst2005tree, riedmiller2005neural}. Recently,  \citet{agarwal2019optimistic}, \citet{fujimoto2019off}, \citet{kumar2019stabilizing} and \citet{siegel2020keep} have proposed new offline-RL algorithms and shown that they outperform off-the-shelf off-policy RL methods. There have also been several new methods explicitly addressing the issues stemming from extrapolation error \citep{fujimoto2018off}. Imitation learning algorithms such as R2D3 \citep{paine2019making} relate to the offline RL algorithms called growing-batch RL algorithms which grows the offline datasets with environment interactions. In this paper, we use Critic Regularized Regression (CRR; \citet{wang2020critic}) as our main offline RL method; CRR uses the advantage function to select the best actions in the dataset for training behavior cloning.

Hyperparameter tuning has been an essential tool for improving the performance of algorithms on many tasks and domains \citep{bergstra2012random, li2017hyperband, snoek2012practical, jaderberg2017population}. \cite{bergstra2012random} have shown that a simple heuristic based on random search can improve the performance of the deep supervised learning models. \cite{snoek2012practical} proposed a Bayesian optimization-based approach and showed that automatic hyperparameter tuning with different metrics can outperform grid and random search. \cite{jaderberg2017population} proposed a population-based evolutionary algorithm and showed promising results on challenging problems such as machine translation and speech synthesis. \cite{melis2017state} showed that a carefully hyperparameter-tuned LSTM model can outperform much more complicated state of art language models. 

Offline hyperparameter tuning for offline RL has received relatively little attention. \cite{farahmand2011model} have proposed BerMin for offline model selection, and they investigated the convergence and other theoretical properties of it. \cite{irpan2019off} proposed Off-Policy Classification (OPC) and evaluate policies with rank correlation on goal-directed continuous control tasks. More recently, \cite{gulcehre2020rl} and \cite{fu2020d4rl} have proposed protocols for evaluating offline RL methods that involve evaluation on a separate set of tasks that are not available for hyperparameter tuning. This type of approach is useful for assessing how well offline RL methods can do without the ability to tune hyperparameters on the target domain.

The early OPE methods relied on importance sampling \citep{precup2000eligibility, precup2001off}. Importance sampling OPE methods can suffer from prohibitively large variance. One solution is to normalize the importance sampling corrections; another is to cap the maximum value of the importance sampling corrections \citep{munos2016safe}. However, the dependence on the ratio of action-selection probabilities makes all these methods challenging to apply in continuous action spaces. More recently, marginal importance sampling methods have been developed that directly estimate the stationary state distribution ratio under the two policies \citep{liu2018breaking, nachum2019dualdice}. The method proposed in \cite{liu2018breaking} require knowledge of the behavior policy. Subsequent works (e.g. \cite{nachum2019dualdice, uehara2019minimax}) require optimization of bilinear minimax objectives that is potentially difficult without effective regularization.

Another class of approaches to the OPE problem is model-based methods. These first estimate a transition model from the offline data, and then perform policy evaluation in the MDP defined by the transition model \cite{mannor2004bias}. The main issue with model-based OPE is that they can introduce a large amount of bias due to the true dynamics not being accurately captured by the learned model, which is often the case for pixel-based control. Moreover, OPE methods using doubly robust estimators combine both model-based and importance sampling based approaches \citep{dudik2011doubly, jiang2016doubly, thomas2016data, mrdr}. However, doubly robust methods are unlikely to be helpful when neither IS nor MB methods perform well, as is likely to be the case in pixel-based continuous control domains.

One approach to off-policy evaluation that avoids both importance sampling corrections and learning transition models is to only perform the policy evaluation step of a policy improvement algorithm. This can be done for instance by using the evaluation policy to select the action $a'$ in the value function $Q(s', a')$ used in most TD-style RL algorithms. Although this is a conceptually very simple approach, the earliest explicit reference to using it for OPE seems to be from \citet{le2019batch}, who refer to it as Fitted Q Evaluation (FQE). It can be seen as a special case of Retrace \citep{munos2016safe} with $T=0$ and $c=1$.

A comprehensive empirical study of OPE methods was carried out in \citet{voloshin2019empirical}. The study considered most of the aforementioned OPE methods and finds FQE to be surprisingly effective despite its simplicity. 
Compared to this work, \citet{voloshin2019empirical} examine simpler environments and datasets. The study also does not address model selection directly and mostly focuses on MSE as the evaluation metric.
\citet{farahmand2011model} consider the problem of model selection for RL. Their method, however, resolves around the Bellman error which does not correlate well with policy performance in complicated domains~\citep{irpan2019off}.

\section{Future Work}
This paper focuses on a single OPE method, FQE. We made this choice because FQE only estimates the value function, and we expected this would make it more likely to scale to the types of problems we are interested in. We look forward to seeing importance-sampling, model-based, or other OPE methods scale to these challenging problems.

An important remaining challenge is how to choose hyperparameters for FQE. Although we do show that it is relatively robust to one (number of learner steps), in general this is an open problem. We do not investigate how to determine if you have sufficient data for reliable hyperparameter selection, or how the type of policy that generated the data affects the quality of the various approaches. Finally, our experiments are performed only in simulation, because it is difficult to get the ground-truth values for a large number of policies learned on robots or other physical systems.

\section{Conclusions}

We provide evidence that by carefully considering the choice of offline RL algorithm, Q estimator, and statistic, we can achieve a strong strategy for offline hyperparameter selection across challenging tasks. In particular, we find that using algorithms that encourage policies to stay close to the behavior policy such as CRR, re-estimating the Q value using FQE, and using $\hat V(s_0)$ as our ranking statistic is sufficient for performing offline hyperparameter selection in the tasks we considered. This is true even in the DM Locomotion tasks, which require control of a 56 degrees of freedom humanoid avatar from visuals provided by an egocentric camera.

\section*{Acknowledgements}

We would like to thank Mohammad Norouzi, Misha Denil, George Tucker and Ofir Nachum for their helpful comments and suggestions.

\bibliographystyle{abbrvnat}
\setlength{\bibsep}{5pt} 
\setlength{\bibhang}{0pt}
\bibliography{main.bib}

\begin{thebibliography}{50}
\providecommand{\natexlab}[1]{#1}
\providecommand{\url}[1]{\texttt{#1}}
\expandafter\ifx\csname urlstyle\endcsname\relax
  \providecommand{\doi}[1]{doi: #1}\else
  \providecommand{\doi}{doi: \begingroup \urlstyle{rm}\Url}\fi

\bibitem[Agarwal et~al.(2019)Agarwal, Schuurmans, and
  Norouzi]{agarwal2019optimistic}
R.~Agarwal, D.~Schuurmans, and M.~Norouzi.
\newblock An optimistic perspective on offline reinforcement learning.
\newblock \emph{arXiv preprint arXiv:1907.04543}, 2019.

\bibitem[Barth-Maron et~al.(2018)Barth-Maron, Hoffman, Budden, Dabney, Horgan,
  Tb, Muldal, Heess, and Lillicrap]{barth2018distributed}
G.~Barth-Maron, M.~W. Hoffman, D.~Budden, W.~Dabney, D.~Horgan, D.~Tb,
  A.~Muldal, N.~Heess, and T.~Lillicrap.
\newblock Distributed distributional deterministic policy gradients.
\newblock \emph{arXiv preprint arXiv:1804.08617}, 2018.

\bibitem[Bellemare et~al.(2017)Bellemare, Dabney, and Munos]{BellemareDM17}
M.~G. Bellemare, W.~Dabney, and R.~Munos.
\newblock A distributional perspective on reinforcement learning.
\newblock \emph{CoRR}, abs/1707.06887, 2017.

\bibitem[Bergstra and Bengio(2012)]{bergstra2012random}
J.~Bergstra and Y.~Bengio.
\newblock Random search for hyper-parameter optimization.
\newblock \emph{The Journal of Machine Learning Research}, 13\penalty0
  (1):\penalty0 281--305, 2012.

\bibitem[Bradtke and Barto(1996)]{bradtke1996linear}
S.~Bradtke and A.~Barto.
\newblock Linear least-squares algorithms for temporal difference learning.
\newblock \emph{Machine Learning}, 22:\penalty0 33--57, 03 1996.

\bibitem[Cabi et~al.(2019)Cabi, Colmenarejo, Novikov, Konyushkova, Reed, Jeong,
  {\.Z}o{\l}na, Aytar, Budden, Vecerik, et~al.]{cabi2019framework}
S.~Cabi, S.~G. Colmenarejo, A.~Novikov, K.~Konyushkova, S.~Reed, R.~Jeong,
  K.~{\.Z}o{\l}na, Y.~Aytar, D.~Budden, M.~Vecerik, et~al.
\newblock A framework for data-driven robotics.
\newblock \emph{arXiv preprint arXiv:1909.12200}, 2019.

\bibitem[Dabney et~al.(2018)Dabney, Ostrovski, Silver, and Munos]{dabney18a}
W.~Dabney, G.~Ostrovski, D.~Silver, and R.~Munos.
\newblock Implicit quantile networks for distributional reinforcement learning.
\newblock In J.~Dy and A.~Krause, editors, \emph{Proceedings of the 35th
  International Conference on Machine Learning}, volume~80 of \emph{Proceedings
  of Machine Learning Research}, pages 1096--1105, Stockholmsmässan, Stockholm
  Sweden, 10--15 Jul 2018. PMLR.

\bibitem[Dud{\'{\i}}k et~al.(2011)Dud{\'{\i}}k, Langford, and
  Li]{dudik2011doubly}
M.~Dud{\'{\i}}k, J.~Langford, and L.~Li.
\newblock Doubly robust policy evaluation and learning.
\newblock \emph{CoRR}, abs/1103.4601, 2011.

\bibitem[Ernst et~al.(2005)Ernst, Geurts, and Wehenkel]{ernst2005tree}
D.~Ernst, P.~Geurts, and L.~Wehenkel.
\newblock Tree-based batch mode reinforcement learning.
\newblock \emph{Journal of Machine Learning Research}, 6:\penalty0 503--556,
  2005.

\bibitem[Farahmand and Szepesv{\'a}ri(2011)]{farahmand2011model}
A.-m. Farahmand and C.~Szepesv{\'a}ri.
\newblock Model selection in reinforcement learning.
\newblock \emph{Machine learning}, 85\penalty0 (3):\penalty0 299--332, 2011.

\bibitem[Farajtabar et~al.(2018)Farajtabar, Chow, and Ghavamzadeh]{mrdr}
M.~Farajtabar, Y.~Chow, and M.~Ghavamzadeh.
\newblock More robust doubly robust off-policy evaluation.
\newblock \emph{CoRR}, abs/1802.03493, 2018.

\bibitem[Fu et~al.(2020)Fu, Kumar, Nachum, Tucker, and Levine]{fu2020d4rl}
J.~Fu, A.~Kumar, O.~Nachum, G.~Tucker, and S.~Levine.
\newblock D4rl: Datasets for deep data-driven reinforcement learning.
\newblock \emph{arXiv preprint arXiv:2004.07219}, 2020.

\bibitem[Fujimoto et~al.(2018)Fujimoto, Meger, and Precup]{fujimoto2018off}
S.~Fujimoto, D.~Meger, and D.~Precup.
\newblock Off-policy deep reinforcement learning without exploration.
\newblock \emph{arXiv preprint arXiv:1812.02900}, 2018.

\bibitem[Fujimoto et~al.(2019)Fujimoto, Meger, and Precup]{fujimoto2019off}
S.~Fujimoto, D.~Meger, and D.~Precup.
\newblock Off-policy deep reinforcement learning without exploration.
\newblock In \emph{International Conference on Machine Learning}, pages
  2052--2062, 2019.

\bibitem[Futoma et~al.(2020)Futoma, Hughes, and Doshi-Velez]{futoma2020popcorn}
J.~Futoma, M.~C. Hughes, and F.~Doshi-Velez.
\newblock Popcorn: Partially observed prediction constrained reinforcement
  learning.
\newblock \emph{arXiv preprint arXiv:2001.04032}, 2020.

\bibitem[Gulcehre et~al.(2020)Gulcehre, Wang, Novikov, Paine, Colmenarejo,
  Zolna, Agarwal, Merel, Mankowitz, Paduraru, Dulac-Arnold, Li, Norouzi,
  Hoffman, Ofir, George, Heess, and de~Freitas]{gulcehre2020rl}
C.~Gulcehre, Z.~Wang, A.~Novikov, T.~L. Paine, S.~G. Colmenarejo, K.~Zolna,
  R.~Agarwal, J.~Merel, D.~Mankowitz, C.~Paduraru, G.~Dulac-Arnold, J.~Li,
  M.~Norouzi, M.~Hoffman, N.~Ofir, T.~George, N.~Heess, and N.~de~Freitas.
\newblock Rl unplugged: Benchmarks for offline reinforcement learning.
\newblock \emph{arXiv preprint arXiv:2006.13888}, 2020.

\bibitem[Henderson et~al.(2018)Henderson, Islam, Bachman, Pineau, Precup, and
  Meger]{henderson2018deep}
P.~Henderson, R.~Islam, P.~Bachman, J.~Pineau, D.~Precup, and D.~Meger.
\newblock Deep reinforcement learning that matters.
\newblock In \emph{Thirty-Second AAAI Conference on Artificial Intelligence},
  2018.

\bibitem[Hoffman et~al.(2020)Hoffman, Shahriari, Aslanides, Barth-Maron,
  Behbahani, Norman, Abdolmaleki, Cassirer, Yang, Baumli, Henderson, Novikov,
  Colmenarejo, Cabi, Gulcehre, Paine, Cowie, Wang, Piot, and
  de~Freitas]{hoffman2020acme}
M.~Hoffman, B.~Shahriari, J.~Aslanides, G.~Barth-Maron, F.~Behbahani,
  T.~Norman, A.~Abdolmaleki, A.~Cassirer, F.~Yang, K.~Baumli, S.~Henderson,
  A.~Novikov, S.~G. Colmenarejo, S.~Cabi, C.~Gulcehre, T.~L. Paine, A.~Cowie,
  Z.~Wang, B.~Piot, and N.~de~Freitas.
\newblock Acme: A research framework for distributed reinforcement learning.
\newblock \emph{arXiv preprint arXiv:2006.00979}, 2020.

\bibitem[Irpan et~al.(2019)Irpan, Rao, Bousmalis, Harris, Ibarz, and
  Levine]{irpan2019off}
A.~Irpan, K.~Rao, K.~Bousmalis, C.~Harris, J.~Ibarz, and S.~Levine.
\newblock Off-policy evaluation via off-policy classification.
\newblock In \emph{Advances in Neural Information Processing Systems}, pages
  5437--5448, 2019.

\bibitem[Jaderberg et~al.(2017)Jaderberg, Dalibard, Osindero, Czarnecki,
  Donahue, Razavi, Vinyals, Green, Dunning, Simonyan,
  et~al.]{jaderberg2017population}
M.~Jaderberg, V.~Dalibard, S.~Osindero, W.~M. Czarnecki, J.~Donahue, A.~Razavi,
  O.~Vinyals, T.~Green, I.~Dunning, K.~Simonyan, et~al.
\newblock Population based training of neural networks.
\newblock \emph{arXiv preprint arXiv:1711.09846}, 2017.

\bibitem[Jaques et~al.(2019)Jaques, Ghandeharioun, Shen, Ferguson, Lapedriza,
  Jones, Gu, and Picard]{jaques2019way}
N.~Jaques, A.~Ghandeharioun, J.~H. Shen, C.~Ferguson, A.~Lapedriza, N.~Jones,
  S.~Gu, and R.~Picard.
\newblock Way off-policy batch deep reinforcement learning of implicit human
  preferences in dialog.
\newblock \emph{arXiv preprint arXiv:1907.00456}, 2019.

\bibitem[Jiang and Li(2016)]{jiang2016doubly}
N.~Jiang and L.~Li.
\newblock Doubly robust off-policy value evaluation for reinforcement learning.
\newblock In M.~F. Balcan and K.~Q. Weinberger, editors, \emph{Proceedings of
  The 33rd International Conference on Machine Learning}, volume~48 of
  \emph{Proceedings of Machine Learning Research}, pages 652--661, 2016.

\bibitem[Kumar et~al.(2019)Kumar, Fu, Soh, Tucker, and
  Levine]{kumar2019stabilizing}
A.~Kumar, J.~Fu, M.~Soh, G.~Tucker, and S.~Levine.
\newblock Stabilizing off-policy q-learning via bootstrapping error reduction.
\newblock In \emph{Advances in Neural Information Processing Systems}, pages
  11761--11771, 2019.

\bibitem[Lagoudakis and Parr(2003)]{lagoudakis2003least}
M.~G. Lagoudakis and R.~Parr.
\newblock Least-squares policy iteration.
\newblock \emph{Journal of Machine Learning Research}, 4:\penalty0 1107--1149,
  2003.

\bibitem[Le et~al.(2019)Le, Voloshin, and Yue]{le2019batch}
H.~Le, C.~Voloshin, and Y.~Yue.
\newblock Batch policy learning under constraints.
\newblock In K.~Chaudhuri and R.~Salakhutdinov, editors, \emph{Proceedings of
  the 36th International Conference on Machine Learning}, volume~97 of
  \emph{Proceedings of Machine Learning Research}, pages 3703--3712, 2019.

\bibitem[Levine et~al.(2020)Levine, Kumar, Tucker, and Fu]{levine2020offline}
S.~Levine, A.~Kumar, G.~Tucker, and J.~Fu.
\newblock Offline reinforcement learning: Tutorial, review, and perspectives on
  open problems.
\newblock \emph{arXiv preprint arXiv:2005.01643}, 2020.

\bibitem[Li et~al.(2017)Li, Jamieson, DeSalvo, Rostamizadeh, and
  Talwalkar]{li2017hyperband}
L.~Li, K.~Jamieson, G.~DeSalvo, A.~Rostamizadeh, and A.~Talwalkar.
\newblock Hyperband: A novel bandit-based approach to hyperparameter
  optimization.
\newblock \emph{The Journal of Machine Learning Research}, 18\penalty0
  (1):\penalty0 6765--6816, 2017.

\bibitem[Liu et~al.(2018)Liu, Li, Tang, and Zhou]{liu2018breaking}
Q.~Liu, L.~Li, Z.~Tang, and D.~Zhou.
\newblock Breaking the curse of horizon: Infinite-horizon off-policy
  estimation.
\newblock In \emph{Advances in Neural Information Processing Systems}. 2018.

\bibitem[Mannor et~al.(2004)Mannor, Simester, Sun, and
  Tsitsiklis]{mannor2004bias}
S.~Mannor, D.~Simester, P.~Sun, and J.~N. Tsitsiklis.
\newblock Bias and variance in value function estimation.
\newblock In \emph{Proceedings of the twenty-first international conference on
  Machine learning}, page~72, 2004.

\bibitem[Melis et~al.(2017)Melis, Dyer, and Blunsom]{melis2017state}
G.~Melis, C.~Dyer, and P.~Blunsom.
\newblock On the state of the art of evaluation in neural language models.
\newblock \emph{arXiv preprint arXiv:1707.05589}, 2017.

\bibitem[Merel et~al.(2018)Merel, Hasenclever, Galashov, Ahuja, Pham, Wayne,
  Teh, and Heess]{merel2018neural}
J.~Merel, L.~Hasenclever, A.~Galashov, A.~Ahuja, V.~Pham, G.~Wayne, Y.~W. Teh,
  and N.~Heess.
\newblock Neural probabilistic motor primitives for humanoid control.
\newblock \emph{arXiv preprint arXiv:1811.11711}, 2018.

\bibitem[Munos et~al.(2016)Munos, Stepleton, Harutyunyan, and
  Bellemare]{munos2016safe}
R.~Munos, T.~Stepleton, A.~Harutyunyan, and M.~G. Bellemare.
\newblock Safe and efficient off-policy reinforcement learning.
\newblock \emph{arXiv preprint arXiv:1606.02647}, 2016.

\bibitem[Nachum et~al.(2019)Nachum, Chow, Dai, and Li]{nachum2019dualdice}
O.~Nachum, Y.~Chow, B.~Dai, and L.~Li.
\newblock Dualdice: Behavior-agnostic estimation of discounted stationary
  distribution corrections.
\newblock In \emph{Advances in Neural Information Processing Systems}. 2019.

\bibitem[Paine et~al.(2019)Paine, Gulcehre, Shahriari, Denil, Hoffman, Soyer,
  Tanburn, Kapturowski, Rabinowitz, Williams, et~al.]{paine2019making}
T.~L. Paine, C.~Gulcehre, B.~Shahriari, M.~Denil, M.~Hoffman, H.~Soyer,
  R.~Tanburn, S.~Kapturowski, N.~Rabinowitz, D.~Williams, et~al.
\newblock Making efficient use of demonstrations to solve hard exploration
  problems.
\newblock \emph{arXiv preprint arXiv:1909.01387}, 2019.

\bibitem[Pomerleau(1989)]{pomerleau1989alvinn}
D.~A. Pomerleau.
\newblock Alvinn: An autonomous land vehicle in a neural network.
\newblock In \emph{Advances in neural information processing systems}, pages
  305--313, 1989.

\bibitem[Precup(2000)]{precup2000eligibility}
D.~Precup.
\newblock Eligibility traces for off-policy policy evaluation.
\newblock \emph{Computer Science Department Faculty Publication Series},
  page~80, 2000.

\bibitem[Precup et~al.(2001)Precup, Sutton, and Dasgupta]{precup2001off}
D.~Precup, R.~S. Sutton, and S.~Dasgupta.
\newblock Off-policy temporal-difference learning with function approximation.
\newblock In \emph{ICML}, pages 417--424, 2001.

\bibitem[Riedmiller(2005)]{riedmiller2005neural}
M.~Riedmiller.
\newblock Neural fitted {Q} iteration -- first experiences with a data
  efficient neural reinforcement learning method.
\newblock In J.~Gama, R.~Camacho, P.~B. Brazdil, A.~M. Jorge, and L.~Torgo,
  editors, \emph{European Conference on Machine Learning}, pages 317--328,
  2005.

\bibitem[Siegel et~al.(2020)Siegel, Springenberg, Berkenkamp, Abdolmaleki,
  Neunert, Lampe, Hafner, Heess, and Riedmiller]{siegel2020keep}
N.~Siegel, J.~T. Springenberg, F.~Berkenkamp, A.~Abdolmaleki, M.~Neunert,
  T.~Lampe, R.~Hafner, N.~Heess, and M.~Riedmiller.
\newblock Keep doing what worked: Behavior modelling priors for offline
  reinforcement learning.
\newblock In \emph{International Conference on Learning Representations}, 2020.

\bibitem[Snoek et~al.(2012)Snoek, Larochelle, and Adams]{snoek2012practical}
J.~Snoek, H.~Larochelle, and R.~P. Adams.
\newblock Practical bayesian optimization of machine learning algorithms.
\newblock In \emph{Advances in neural information processing systems}, pages
  2951--2959, 2012.

\bibitem[Sutton and Barto(1998)]{sutton1998}
R.~S. Sutton and A.~G. Barto.
\newblock Reinforcement learning: An introduction.
\newblock \emph{{IEEE} Trans. Neural Networks}, 1998.

\bibitem[Tassa et~al.(2018)Tassa, Doron, Muldal, Erez, Li, Casas, Budden,
  Abdolmaleki, Merel, Lefrancq, Lillicrap, and Riedmiller]{tassa2018deepmind}
Y.~Tassa, Y.~Doron, A.~Muldal, T.~Erez, Y.~Li, D.~d.~L. Casas, D.~Budden,
  A.~Abdolmaleki, J.~Merel, A.~Lefrancq, T.~Lillicrap, and M.~Riedmiller.
\newblock Deepmind control suite.
\newblock \emph{arXiv preprint arXiv:1801.00690}, 2018.

\bibitem[Tassa et~al.(2020)Tassa, Tunyasuvunakool, Muldal, Doron, Liu, Bohez,
  Merel, Erez, Lillicrap, and Heess]{tassa2020dm_control}
Y.~Tassa, S.~Tunyasuvunakool, A.~Muldal, Y.~Doron, S.~Liu, S.~Bohez, J.~Merel,
  T.~Erez, T.~Lillicrap, and N.~Heess.
\newblock dm{\_}control: Software and tasks for continuous control.
\newblock \emph{arXiv preprint arXiv:2006.12983}, 2020.

\bibitem[Thomas and Brunskill(2016)]{thomas2016data}
P.~Thomas and E.~Brunskill.
\newblock Data-efficient off-policy policy evaluation for reinforcement
  learning.
\newblock In M.~F. Balcan and K.~Q. Weinberger, editors, \emph{Proceedings of
  The 33rd International Conference on Machine Learning}, volume~48 of
  \emph{Proceedings of Machine Learning Research}, pages 2139--2148, 2016.

\bibitem[Todorov et~al.(2012)Todorov, Erez, and Tassa]{todorov2012mujoco}
E.~Todorov, T.~Erez, and Y.~Tassa.
\newblock Mujoco: A physics engine for model-based control.
\newblock In \emph{2012 IEEE/RSJ International Conference on Intelligent Robots
  and Systems}, 2012.

\bibitem[Uehara and Jiang(2019)]{uehara2019minimax}
M.~Uehara and N.~Jiang.
\newblock Minimax weight and q-function learning for off-policy evaluation.
\newblock \emph{arXiv preprint arXiv:1910.12809}, 2019.

\bibitem[Vecerik et~al.(2017)Vecerik, Hester, Scholz, Wang, Pietquin, Piot,
  Heess, Roth{\"o}rl, Lampe, and Riedmiller]{vecerik2017leveraging}
M.~Vecerik, T.~Hester, J.~Scholz, F.~Wang, O.~Pietquin, B.~Piot, N.~Heess,
  T.~Roth{\"o}rl, T.~Lampe, and M.~Riedmiller.
\newblock Leveraging demonstrations for deep reinforcement learning on robotics
  problems with sparse rewards.
\newblock \emph{arXiv preprint arXiv:1707.08817}, 2017.

\bibitem[Voloshin et~al.(2019)Voloshin, Le, Jiang, and
  Yue]{voloshin2019empirical}
C.~Voloshin, H.~M. Le, N.~Jiang, and Y.~Yue.
\newblock Empirical study of off-policy policy evaluation for reinforcement
  learning.
\newblock \emph{arXiv preprint arXiv:1911.06854}, 2019.

\bibitem[Wang et~al.(2020)Wang, Novikov, Żołna, Springenberg, Reed,
  Shahriari, Siegel, Merel, Gulcehre, Heess, and de~Freitas]{wang2020critic}
Z.~Wang, A.~Novikov, K.~Żołna, J.~T. Springenberg, S.~Reed, B.~Shahriari,
  N.~Siegel, J.~Merel, C.~Gulcehre, N.~Heess, and N.~de~Freitas.
\newblock Critic regularized regression.
\newblock \emph{arXiv preprint arXiv:2006.15134}, 2020.

\bibitem[Wu et~al.(2019)Wu, Tucker, and Nachum]{wu2019behavior}
Y.~Wu, G.~Tucker, and O.~Nachum.
\newblock Behavior regularized offline reinforcement learning.
\newblock \emph{arXiv preprint arXiv:1911.11361}, 2019.

\end{thebibliography}

\clearpage
\appendix
\begin{center}
{\Large \bf Appendix}
\end{center}
\normalsize

\section{Absolute Error Results}
In the main paper we provide over-estimation results, summarized for each task domain. In this section we show the absolute error for each task individually.

\begin{figure}[h!]
\includegraphics[width=\linewidth]{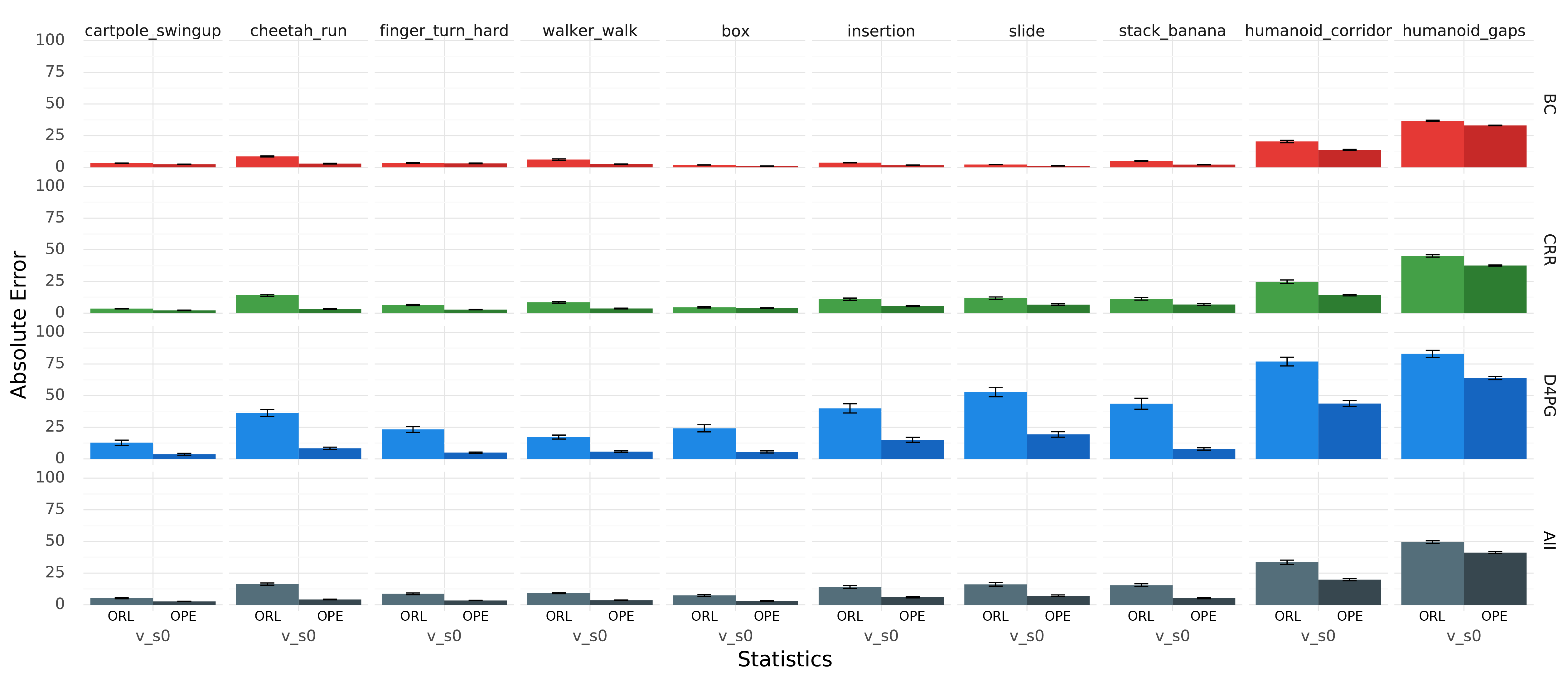}
\caption{\textbf{Absolute Error.} We compute the absolute error between the actual values and V(s0) statistics, using the ORL and OPE critics. We find two main error trends: first in terms of algorithms, statistics tend to have highest error on D4PG, followed by CRR, followed by BC. Second in terms of task domains,  statistics tend to have highest error on DM Locomotion, followed by Manipulation Playground, followed by DM Control Suite. And OPE statistics have lower error and ORL statistics.}
\label{fig:absolute_error}
\end{figure}

\clearpage
\section{Additional Ranking Results}
\label{app:ranking_results}
In addition to $\hat V(s_0)$ and Soft OPC, we also considered two additional offline statstics: 

\begin{itemize}
    \item \textbf{avg\_q} Use the expected value across all states in the dataset $\mathbb{E}_{s\sim {\mathcal D}}[Q_\theta(s, \pi(s))]$. This differs slightly from what we care about, it corresponds to running the policy from any state along a trajectory in the behavior data. Nevertheless this statistic performs similarly to $\hat V(s_0)$.
    \item \textbf{td\_err} Use the average temporal difference error across all (s,a,r,s') tuples in dataset. \\$\mathbb{E}_{(s,a,r,s')\sim {\mathcal D}}[r + \gamma Q_\theta(s', \pi(s')) - Q_\theta(s, a)]$. This statistic is more indicative of the quality of the critic than the quality of the policy, which may explain why it performs quite poorly for our purposes.
\end{itemize}

\begin{figure}[h!]
\includegraphics[width=\linewidth]{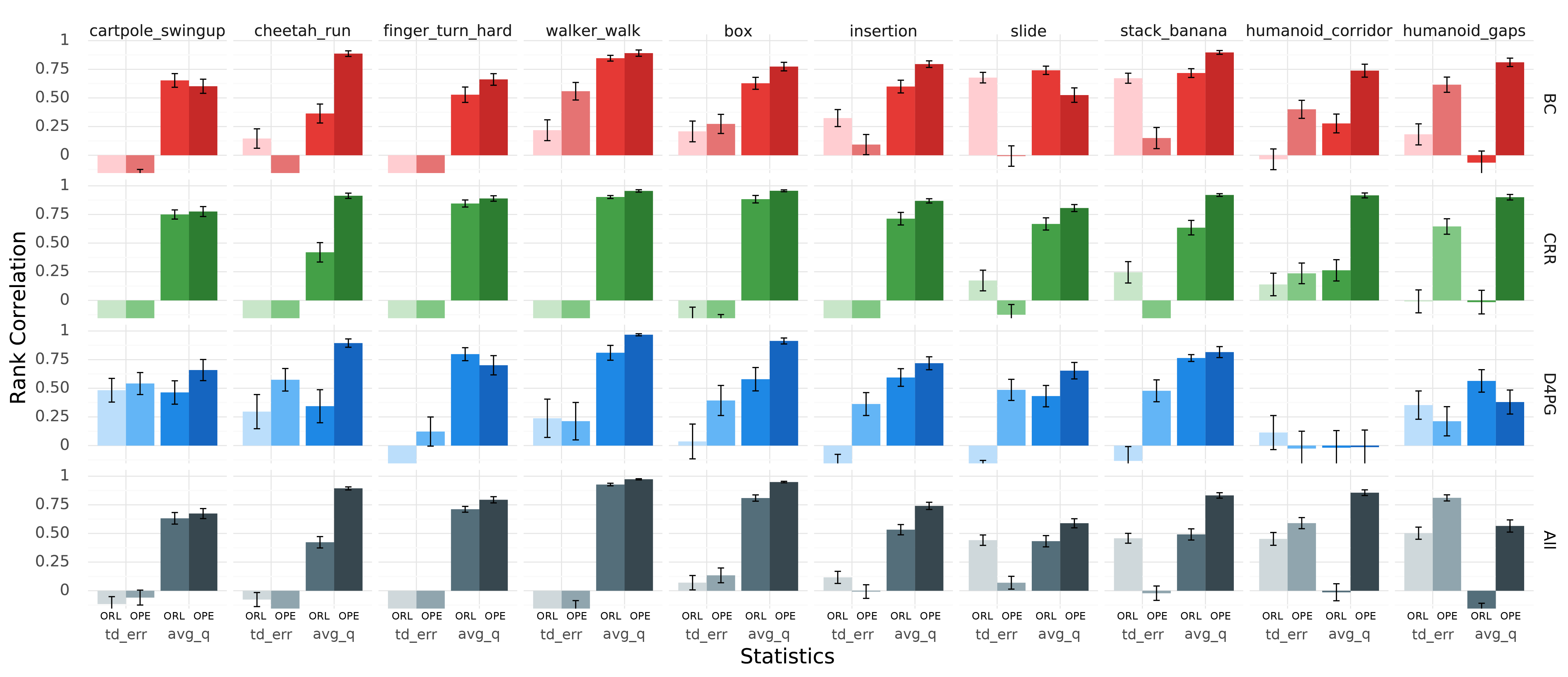}
\caption{\textbf{Rank correlation within algorithm and across all algorithms cont'd.} We compare the rank correlation between the actual value and additional policy statistics from ORL and OPE critics (avg\_q, td\_err). In general, avg\_q follows similar trends to $\hat V(s_0)$, but its slightly worse, and td\_err performs quite poorly overall.}
\label{fig:rank_correlation_appendix}
\end{figure}

\begin{figure}[h!]
\includegraphics[width=\linewidth]{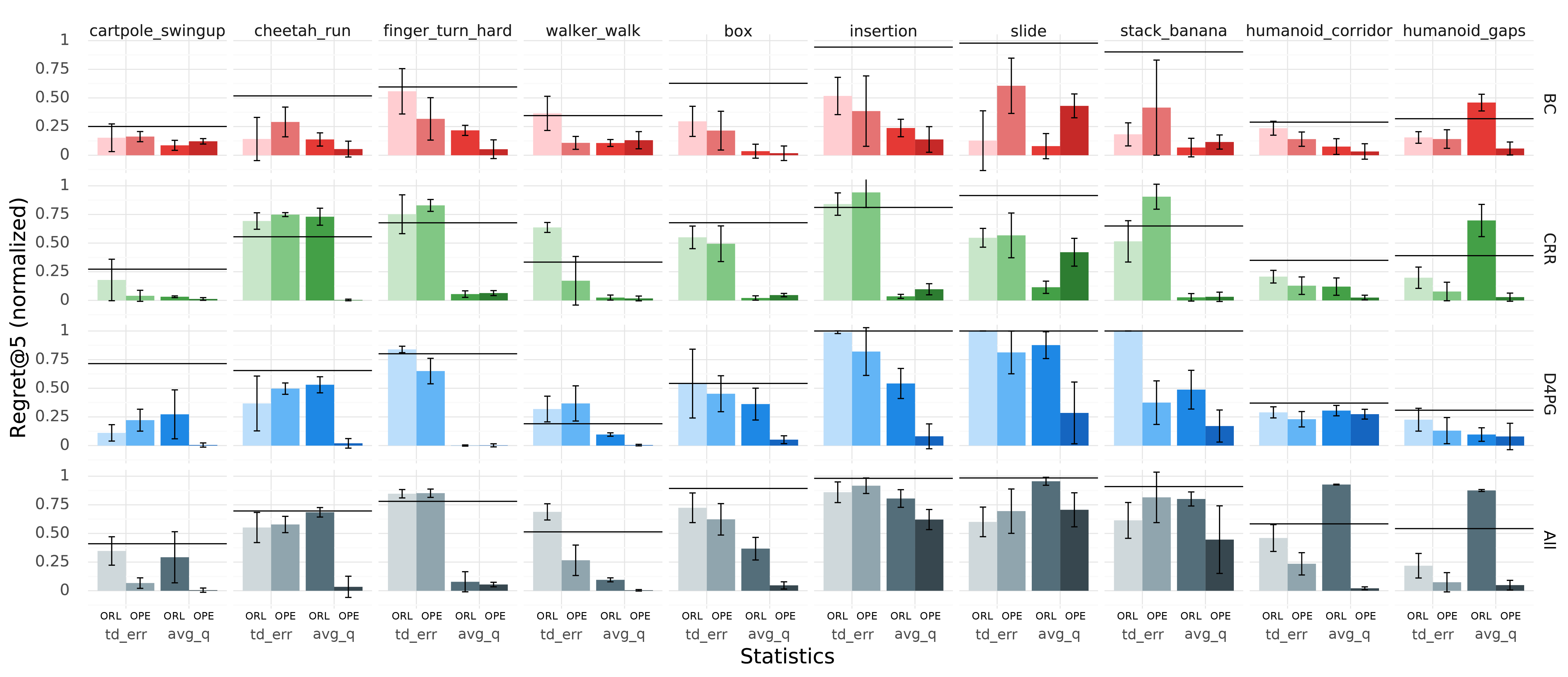}
\caption{\textbf{Regret@5 within algorithm and across all algorithms cont'd.} We compare the normalized regret@5 between the actual value and additional policy statistics from ORL and OPE critics (avg\_q, td\_err). The regret follows similar trends to rank correlation. In general, avg\_q follows similar trends to $\hat V(s_0)$, but its slightly worse, and td\_err performs quite poorly overall.}
\label{fig:regret_appendix}
\end{figure}

\clearpage
\section{Fitted Q Evaluation without Distributional Critic}
\label{app:fqe_without_distributional}
We aimed to keep the critic loss consistent for all experiments in the main paper. But some readers may be curious how FQE would perform without a distributional critic. We re-ran our FQE evaluation with and without a distributional critic on the DM Control Suite tasks.

\begin{figure}[h!]
\includegraphics[width=\linewidth]{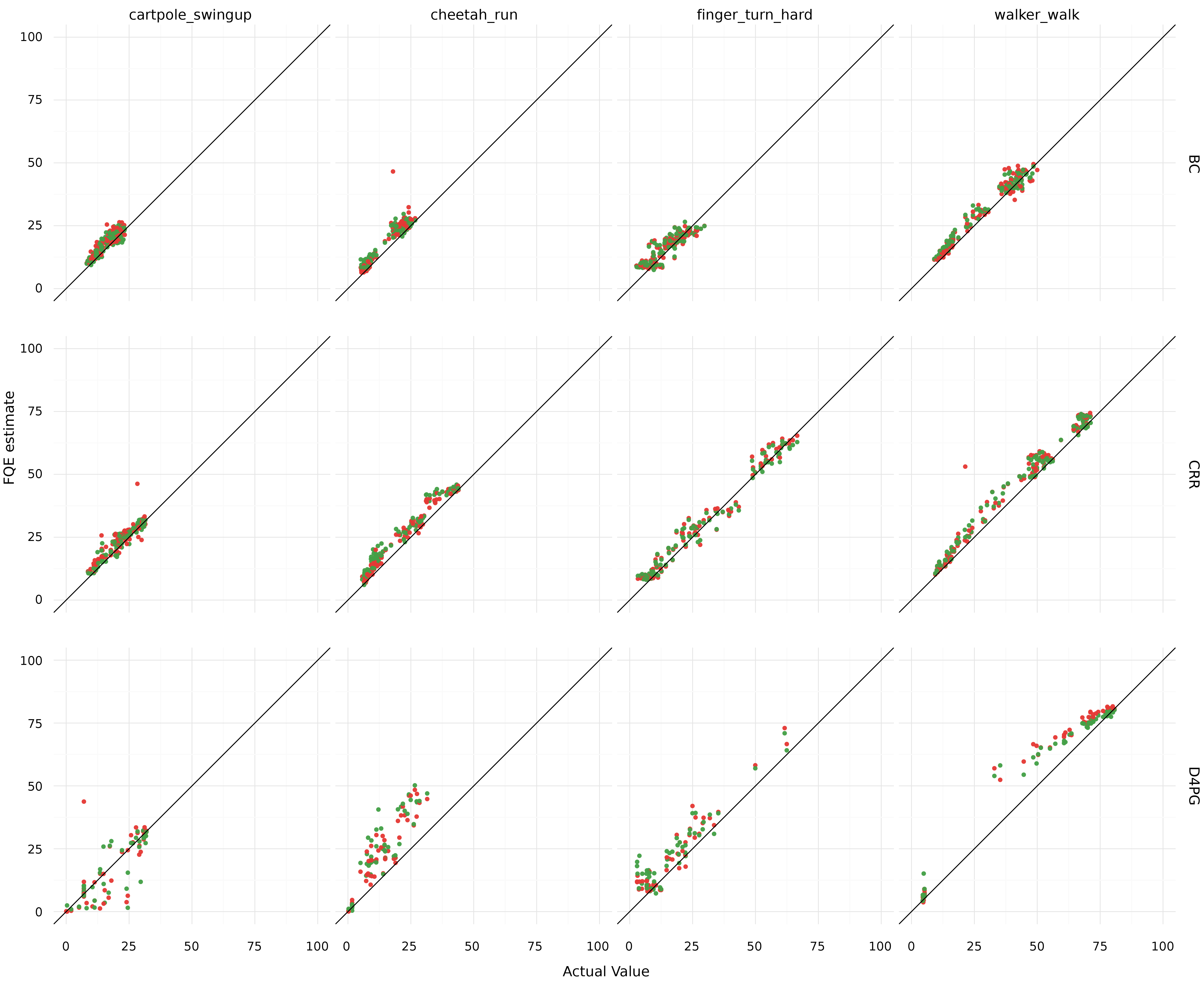}
\caption{\textbf{Comparing FQE estimates with and without distributional critics.} (\textcolor{plot_red}{$\blacksquare$} With, \textcolor{plot_green}{$\blacksquare$} Without) Overall estimates fall within a similar range. FQE with a distributional critic has a few outliers with high values that FQE without does not. On closer inspection we found these corresponded to experiments that were terminated early.}
\label{fig:XXX}
\end{figure}

\clearpage
\section{Fitted Q Evaluation Code}
\label{sec:fqe_code}

We wrote our FQE code using TensorFlow 2 and Acme \citep{hoffman2020acme}. Listing \ref{lst:fqe} is a simplified version of the code we used.

\begin{figure}[h!]
\begin{minipage}{\linewidth}
\begin{lstlisting}[caption={\textbf{Simplified code for of Fitted Q Evaluation.} The code is functional despite its simplicity.},label={lst:fqe},language=Python]
# Copyright 2020 DeepMind Technologies Limited.
# SPDX-License-Identifier: Apache-2.0

discount = 0.99
target_update_period = 100
 
num_steps = 0
 
for o_tm1, a_tm1, d_t, is_terminal, o_t in dataset:
  q_t = target_critic_network(o_t, policy_network(o_t))
 
  # Operations that will be differentiated should be executed in
  # this context.
  with tf.GradientTape() as tape:
    q_tm1 = critic_network(o_tm1, a_tm1)
 
    # Use 0 discount at terminal states.
    curr_discount = 0.0 if is_terminal else discount
    critic_loss = losses.categorical(q_tm1, r_t, curr_discount, q_t)
 
  # Get trainable variables.
  variables = critic_network.trainable_variables
 
  # Compute gradients.
  gradients = tape.gradient(critic_loss, variables)
 
  # Apply gradients.
  optimizer.apply(gradients, variables)
 
  # Update online -> target parameters if necessary.
  source_variables = critic_network.variables
  target_variables = target_critic_network.variables
  if num_steps % target_update_period == 0:
    for src, dest in zip(source_variables, target_variables):
      dest.assign(src)
  num_steps += 1
\end{lstlisting}
\end{minipage}
\end{figure}

\clearpage
\section{Dataset Details}
\label{app:dataset}
\begin{wrapfigure}{r}{0.45\linewidth}
\vspace{-0.475cm}
\centering
\refstepcounter{table}
\caption*{Table \thetable~~| \textbf{Dataset sizes in number of episodes.} \label{tab:data_sizes}}
\small
\begin{tabular}{l|r}
\toprule
Task & No. Episodes \\
\hline
Cartpole Swingup & 100 \\
Cheetah Run & 2000 \\
Finger Turn Hard & 2000 \\
Walker Walk & 800 \\
\hline
Box & 8000 \\
Stack Banana & 8000 \\
Insertion & 8000 \\
Slide & 8000 \\
\hline
Humanoid Corridor & 16000 \\
Humanoid Gaps & 16000 \\
\bottomrule
\end{tabular}
\vspace{-0.45cm}
\end{wrapfigure}
In this section, we provide details regarding the datasets used in this paper. 
For the sizes of all datasets used, please refer to Table~\ref{tab:data_sizes}.

\textbf{DM Control Suite} \hspace{0.2cm}
We largely follow the procedures of generating data as described in~\cite{gulcehre2020rl}. All datasets used in the DM control suite domains are generated by $3$ independent runs of a D4PG agent. Episodes from the entire training run is saved to increase diversity.
Unlike in~\cite{gulcehre2020rl}, we do not further filter out successful episodes and the size of the datasets used in this paper is larger than that in~\cite{gulcehre2020rl}.

\textbf{Manipulation tasks} \hspace{0.2cm}
These datasets are the same as the robotics datasets used in \cite{wang2020critic}. 
The dataset for each task is generated from 3 independent runs
of a D4PGfD agent\footnote{D4PGfD is similar to DDPGfD \citep{vecerik2017leveraging} but is augmented with distributional critics.} where 100 human demonstrations are used for each task to assist with exploration. 
The dataset contains 8000 episodes from the entire training process and thus consists
of both successful and unsuccessful episodes.

\textbf{DM Locomotion} \hspace{0.2cm}
We largely adhere to the procedures of generating data as described in~\cite{gulcehre2020rl}. For each task, three policies are trained following \cite{merel2018neural}.
Episodes from the entire training runs are saved and sub-sampled to include both successful and failed episodes. 
Unlike in~\cite{gulcehre2020rl}, we do not further filter out successful episodes and the size of the datasets used in this paper is larger than that in~\cite{gulcehre2020rl}.

\end{document}